\documentclass[11pt]{article}

\usepackage[preprint]{latex/acl}

\usepackage{times}
\usepackage{latexsym}

\usepackage[T1]{fontenc}

\usepackage[utf8]{inputenc}

\usepackage{microtype}

\usepackage{inconsolata}

\usepackage{graphicx}
\usepackage{subcaption}

\title{Why Low-Resource NLP Needs More Than Cross-Lingual Transfer: Lessons Learned from Luxembourgish}

\author{
 \textbf{Fred Philippy\textsuperscript{1}},
 \textbf{Siwen Guo\textsuperscript{2}},
 \textbf{Jacques Klein\textsuperscript{1}},
 \textbf{Tegawendé F. Bissyandé\textsuperscript{1}}
\\
\\
 \textsuperscript{1}SnT, University of Luxembourg, Luxembourg
 \\
 \textsuperscript{2}Luxembourg Institute of Science and Technology, Luxembourg
\\
 \small{
   \textbf{Correspondence:} \href{mailto:fred.philippy@uni.lu}{fred.philippy@uni.lu}
 }
}

\begin{document}
\maketitle

\begin{abstract}
Cross-lingual transfer has become a central paradigm for extending natural language processing (NLP) technologies to low-resource languages. By leveraging supervision from high-resource languages, multilingual language models can achieve strong task performance with little or no labeled target-language data. However, it remains unclear to what extent cross-lingual transfer can substitute for language-specific efforts.
In this paper, we synthesize prior research findings and data collection results on Luxembourgish, which, despite its typological proximity to high-resource languages and its presence in a multilingual context, remains insufficiently represented in modern NLP technologies. 
Across findings, we observe a fundamental interdependence between cross-lingual transfer and language-specific efforts. Cross-lingual transfer can substantially improve target-language performance, but its success depends critically on the availability of sufficiently high-quality, task-aligned target-language data. At the same time, such resources, particularly in low-resource settings, are typically too limited in scale to drive strong performance on their own. Instead, such resources reach their full potential only when leveraged within a cross-lingual framework.
We therefore argue that cross-lingual transfer and language-specific efforts should not be viewed as competing alternatives. Instead, they function as complementary components of a sustainable low-resource NLP pipeline. Based on these insights, we provide practical guidelines for integrating and balancing cross-lingual transfer with language-specific development in sustainable low-resource NLP pipelines.
\end{abstract}
\section{Introduction}
Recent advances in multilingual language models have dramatically improved the feasibility of developing NLP systems for low-resource languages. Cross-lingual transfer, where supervision in a high-resource language is leveraged to enable performance in other languages, has emerged as a particularly attractive paradigm. In zero-shot and few-shot settings, multilingual models can often achieve competitive performance on downstream tasks in languages with little or no labeled data, reducing the immediate need for costly language-specific annotation \citep{lin-etal-2022-shot}.

For low-resource communities, these developments are transformative, as cross-lingual methods make it possible to bootstrap systems for classification, inference, or retrieval using predominantly English or other high-resource supervision. This development suggests a compelling narrative: perhaps language-specific resources are no longer strictly necessary if transfer is sufficiently strong.

However, this perspective requires nuance. Cross-lingual transfer comes with several well-known limitations. Its effectiveness can vary substantially across language pairs, and even within the same pair it may behave asymmetrically depending on the transfer direction \citep{malkin-etal-2022-balanced}. More broadly, transfer performance is shaped by multiple interacting factors, often resulting in lower-than-expected gains \citep{philippy-etal-2023-towards}.

In this paper, we consolidate lessons learned from a series of studies on Luxembourgish that highlight these issues in a particularly instructive setting. Despite its low-resource status, Luxembourgish occupies a comparatively advantageous position, given its intensely multilingual context and its typological proximity to high-resource neighboring languages, factors that would be expected to support strong cross-lingual transfer. Yet, the performance of existing language models on Luxembourgish remains limited and falls short of what might be expected given these favorable conditions.

\begin{figure}[t]
    \newcommand{\colorline}[1]{%
      \textcolor[HTML]{#1}{\raisebox{0.4ex}{\rule{0.4cm}{1.5pt}}}%
    }
    \centering
    \includegraphics[width=0.98\linewidth]{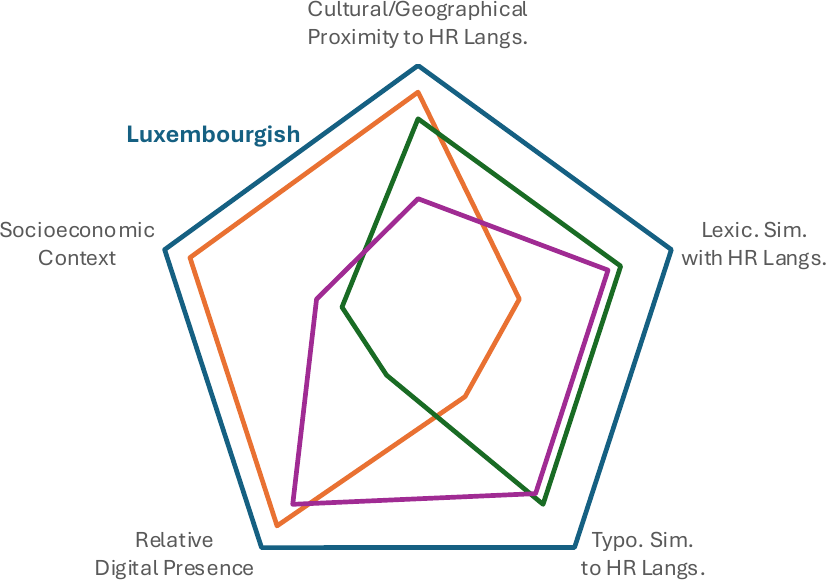}
    \caption{\textbf{Radar chart illustrating structural and sociotechnical dimensions associated with NLP inclusion and cross-lingual transfer: cultural/geographical proximity to high-resource languages, lexical similarity, typological similarity, relative digital presence, and socioeconomic context.}
    Luxembourgish (\colorline{156082}) approximates an upper bound among lower-resource languages, combining structural proximity with strong institutional and socioeconomic support.
    In contrast, many other low-resource languages tend to lack strength along one or more of these axes. Some benefit from favorable socioeconomic conditions and digital capacity but have limited lexical or typological similarity to major high-resource languages (\colorline{E97132}) (e.g., Irish, Basque, Icelandic, Welsh, Finnish, Maltese). Others are structurally and culturally close to at least one high-resource language but have weaker socioeconomic and digital capacity, often due to limited institutionalization or concentration in economically disadvantaged regions (\colorline{196B24}) (e.g., Scots, Galician, Lombard). Others show relatively strong digital presence driven by active online communities despite limited institutional or socioeconomic support, while remaining culturally or geographically distant from most high-resource languages even if structurally related to at least one of them (\colorline{A02B93}) (e.g., Haitian Creole, Nigerian Pidgin).}
    \label{fig:intro}
\end{figure}

At the same time, these structural advantages make Luxembourgish a relative best-case scenario among low-resource languages. Studying where transfer still breaks down in this context allows us to identify core bottlenecks that cannot simply be attributed to extreme data scarcity, and that are therefore likely to generalize beyond this case. In this sense, Luxembourgish may serve as a practical upper bound on what cross-lingual transfer can realistically achieve for low-resource languages (Figure \ref{fig:intro}).

Across findings, we observe that cross-lingual transfer consistently shows strong potential for Luxembourgish, but also shows clear limits to its effectiveness. These constraints often arise from insufficient cross-lingual signal, driven by factors such as low target-language data quality, misalignment between downstream objectives, or unreliable target-language evaluation. In short, cross-lingual transfer is powerful, but not self-sufficient. At the same time, purely language-specific efforts pursued in monolingual isolation are equally limited, as they forgo the benefits of cross-lingual signals.

Therefore, in this paper, we argue that cross-lingual and language-specific efforts are best understood as complementary components of a shared development cycle. Drawing on a series of empirical findings, we illustrate this interplay from several angles. 

Taken together, these findings suggest a nuanced perspective. Cross-lingual transfer is crucial for bootstrapping low-resource NLP, but sustainable performance requires complementary language-specific efforts that ground models in the target language's linguistic and cultural context.

By grounding this discussion in a concrete low-resource case study, we aim to challenge reductive framings that oppose cross-lingual transfer to language-specific development, and instead promote more sustainable and pluralistic research strategies.

\section{The Limits of ``Emergent'' Cross-Lingual Transfer}
Cross-lingual transfer is often presented as an emergent property of multilingual language models \citep{wang-etal-2024-probing-emergence}. Empirical results demonstrate strong transfer performance of multilingual models, often without explicit alignment objectives. The implicit logic is straightforward: once sufficiently diverse multilingual pretraining data are mixed at scale, knowledge is assumed to generalize automatically across languages.
This assumption is also reflected in contemporary model development practices. For instance, the technical report of Qwen-3 \citep{yang2025qwen3technicalreport} describes the models as exhibiting ``improved cross-lingual understanding and generation capabilities'', yet does not report the use of any explicit cross-lingual alignment objectives or dedicated transfer-enhancing mechanisms beyond large-scale multilingual pretraining. The implicit premise is that cross-lingual competence will emerge naturally from joint multilingual training. 

While this assumption is not unfounded, it is incomplete. Although cross-lingual transfer is not an intrinsic objective of most multilingual training regimes, it is a by-product of shared parameterization and distributional overlap. As a consequence, cross-lingual transfer is neither uniform nor guaranteed, and its success depends on a range of factors, including pretraining data composition and lexical or typological proximity between languages \citep{philippy-etal-2023-towards}.
Consequently, languages that are typologically distant, morphologically rich, or underrepresented in web-scale corpora consistently benefit less from the ``emergent'' transfer abilities.

In such cases, transfer performance can often be substantially improved through targeted interventions. Approaches such as supervised alignment techniques \citep{hammerl-etal-2024-understanding}, adapter-based frameworks \citep{pfeiffer-etal-2020-mad, parovic-etal-2022-bad}, or continued pre-training \citep{zheng-etal-2024-breaking, fujii2024continual} have all been shown to boost cross-lingual generalization. While these approaches are valuable and frequently presented as language-agnostic, many implicitly rely on the existence of high-quality resources in the target language. As we show in this paper, this assumption often fails in low-resource contexts, where the scarcity of reliable supervision fundamentally constrains the effectiveness of otherwise promising techniques.

\section{Luxembourgish as a Promising Language for NLP: A Theoretical Perspective}
While Luxembourgish has a relatively small speaker base of approximately 400,000, its structural characteristics and sociolinguistic profile position it (theoretically) as an ideal candidate for inclusion in multilingual NLP.

\textbf{Strong institutional support and standardization.}
Luxembourgish is the official national language of Luxembourg, which ensures formal recognition and sustained government-backed language planning. Public institutions actively maintain linguistic resources, including standardized orthography guidelines and lexicographic databases \citep{cpll_zls_2019_orthografie}. In addition, organizations such as the \textit{Centre for the Luxembourgish Language} provide access to high-quality, curated dictionary resources \citep{zls_lod_2025}\footnote{\url{https://lod.lu}}.

\begin{figure}[h!]
    \centering
    \includegraphics[width=\linewidth]{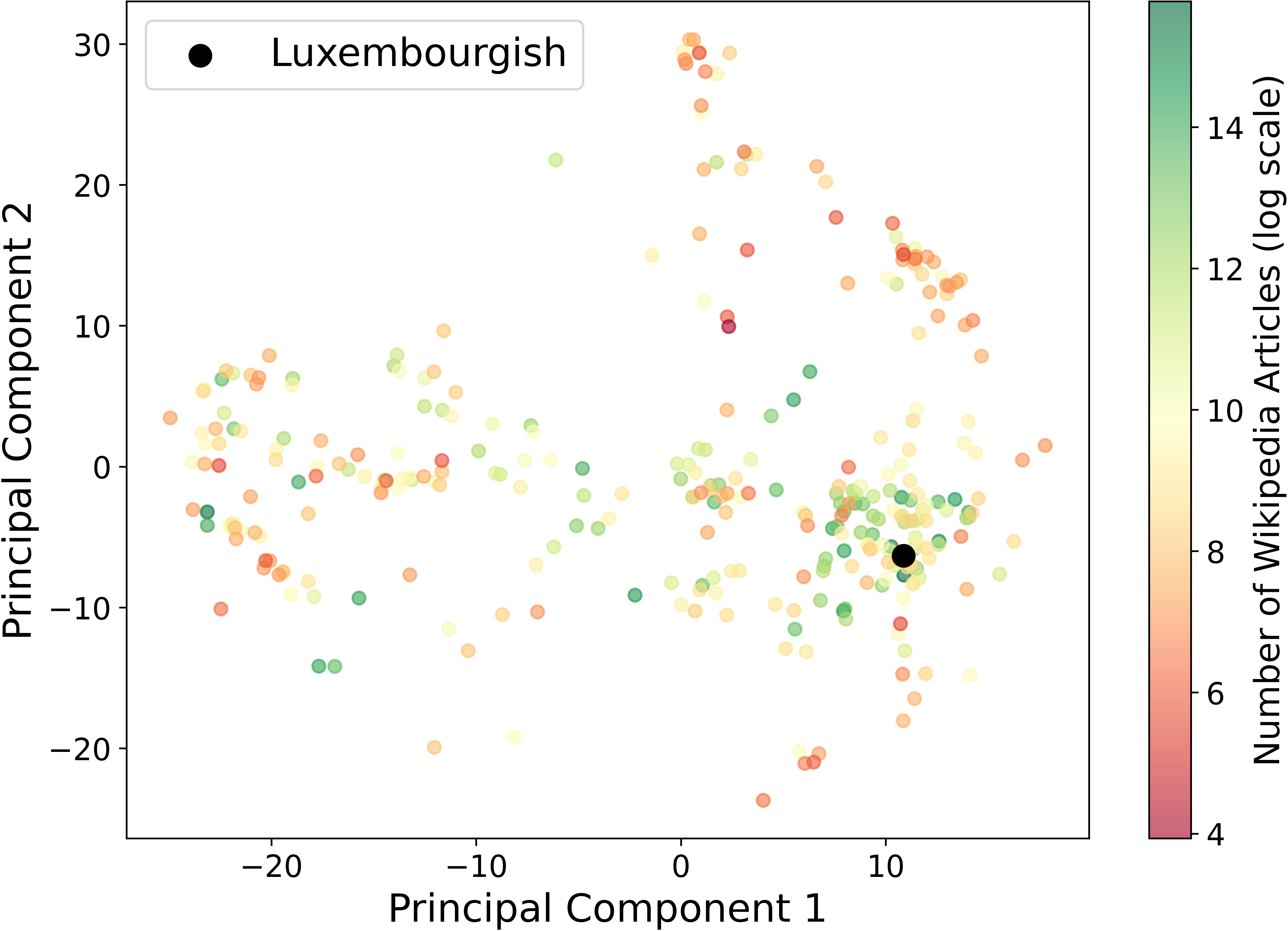}
    \caption{\textbf{PCA projection of concatenated syntactic, phonological, inventory, genetic, and geographical representations for each language.} Each point denotes a language; spatial proximity reflects overall linguistic similarity. Colors indicate the logarithm of the number of Wikipedia articles (resource proxy). Luxembourgish is located within a dense cluster of predominantly mid- to high-resource languages. More details are provided in Appendix \ref{app:lang_dist}.}
    \label{fig:lang_dist}
\end{figure}

\textbf{Favorable conditions for cross-lingual transfer.}
Luxembourgish is well positioned for transfer-based approaches due to its close relationship with German and its extensive lexical borrowing from French. This dual proximity to two major high-resource languages makes it theoretically highly amenable to multilingual modeling and cross-lingual adaptation. Its use of the Latin script further reduces technical barriers related to tokenization, font handling, and script-specific modeling challenges. Overall, Luxembourgish's strong ties to Germanic languages through German and to Romance languages through sustained French influence, embed it firmly within the Indo-European language family and align it closely with many high-resource languages in NLP (Figure \ref{fig:lang_dist}).

\textbf{Cultural proximity to high-resource language settings.}
Luxembourgish is embedded in a broader Western European sociocultural context, with strong ties to both German- and French-speaking regions. This alignment reduces the likelihood of severe distributional mismatches for many common NLP tasks, compared to low-resource languages that are typologically or culturally more distant from the dominant training data of most multilingual language models.

\textbf{A highly multilingual speaker community.}
Luxembourgish is embedded in a highly multilingual speech community \citep{fehlen_linguistic_2023}, in which most native speakers command at least one additional language\footnote{typically French, German, or English}, often at a high or near-native level. This linguistic ecology gives rise to frequent code-switching and cross-lingual interference in both spoken and written communication, providing ecologically valid data for studying mixed-language processing, transfer dynamics, and robustness in multilingual NLP. At the same time, the community’s widespread bilingual and trilingual proficiency, at least in theory, broadens the pool of potential cross-lingual annotators (though generally non-expert), thereby facilitating resource creation for tasks such as machine translation.

\begin{figure}[h!]
    \centering

    \begin{subfigure}{0.99\linewidth}
        \centering
        \includegraphics[width=\linewidth]{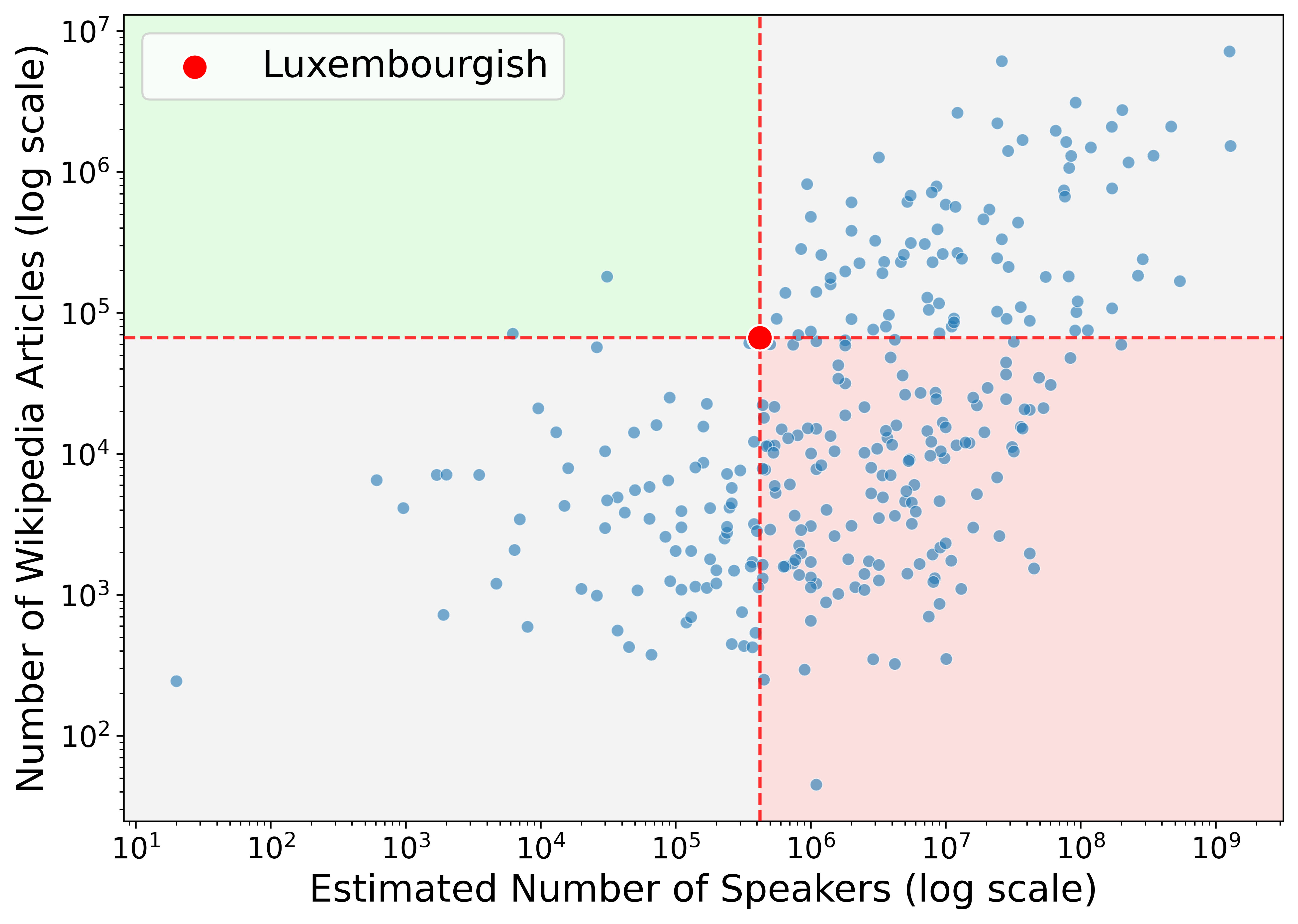}
        \label{fig:speakers_vs_articles_wikipedia}
    \end{subfigure}

    \vspace{0.2em}

    \begin{subfigure}{0.99\linewidth}
        \centering
        \includegraphics[width=\linewidth]{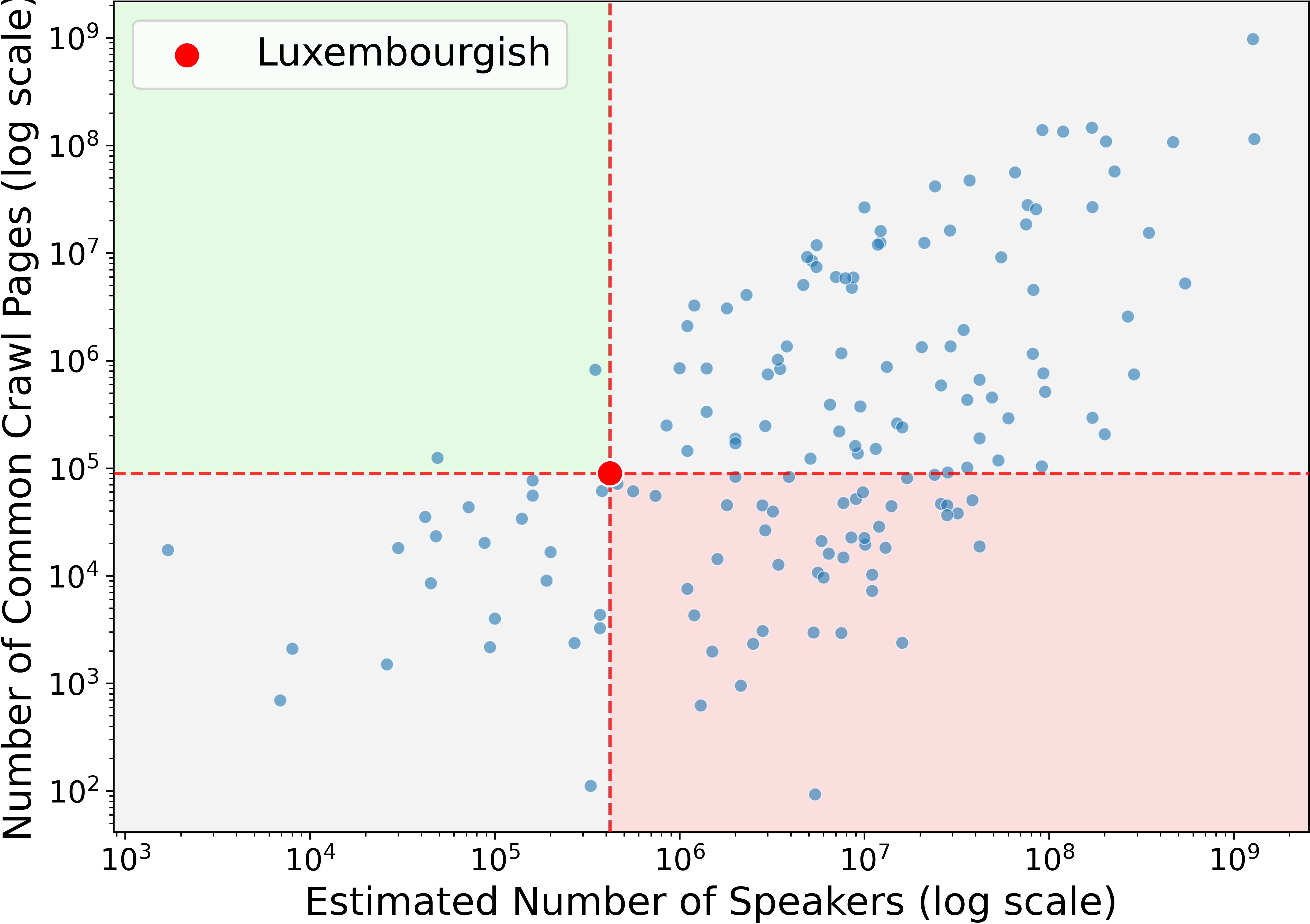}
        \label{fig:speakers_vs_articles_commoncrawl}
    \end{subfigure}

    \caption{\textbf{Estimated number of speakers vs. number of Wikipedia articles (top) / Common Crawl pages (bottom) across languages.} Each point represents a language (both axes shown on a log scale). Shaded quadrants indicate languages with (i) fewer speakers and fewer articles (bottom left), (ii) more speakers but fewer articles (bottom right), (iii) fewer speakers but more articles (top left), and (iv) more speakers and more articles (top right). More details are provided in Appendix \ref{app:speakers_vs_articles}.}
    \label{fig:speakers_vs_articles}
\end{figure}

\textbf{Disproportionately high digital presence for its size.}
Luxembourgish benefits from a relatively rich online ecosystem, including news content\footnote{\url{https://www.rtl.lu}}, its own Wikipedia edition\footnote{\url{https://lb.wikipedia.org/wiki/Haaptsäit}}, and an active multilingual media environment. Despite its small speaker population, Luxembourgish stands out as having unusually high digital coverage relative to its size. Figure~\ref{fig:speakers_vs_articles} illustrates this imbalance: several languages with vastly larger speaker communities are represented by less data\footnote{Languages such as Oromo, Sindhi, Sundanese, Igbo, and Yoruba each have between 50 and 100 times as many speakers as Luxembourgish, yet they are represented by fewer Wikipedia articles and smaller CommonCrawl corpora.}, while almost none with similar or smaller speaker population sizes are better represented.

Despite these advantages, Luxembourgish remains underrepresented in modern NLP systems. It is frequently absent from multilingual model documentation, inconsistently supported in pretraining data, and often excluded from standard multilingual benchmarks. Large closed-source language models remain noticeably less fluent in Luxembourgish than in neighboring high-resource languages. Language confusion, particularly between Luxembourgish and German, is common, and code-switching phenomena are often mishandled.

Moreover, foundational NLP infrastructure for Luxembourgish remains limited. Clean, widely adopted benchmarks are scarce. Core tools such as part-of-speech taggers, dependency parsers, and spellcheckers exist but often lack the robustness and evaluation depth seen for higher-resource languages. For example, the first treebank for Luxembourgish has only recently been created by \citet{plum-etal-2024-luxbank}, but contains to this date merely 20 annotated sentences. In short, Luxembourgish is structurally well-positioned for inclusion in NLP, yet practically under-integrated.

This contrast makes Luxembourgish a particularly informative case study: if cross-lingual transfer were truly automatic, a language with these characteristics should already exhibit strong and stable performance across multilingual systems.

\section{Pushing Transfer in Practice: Lessons from Luxembourgish}
Our empirical work on Luxembourgish shows that cross-lingual transfer can be highly effective, but it is not self-sustaining. In practice, meaningful improvements require deliberate, language-aware interventions rather than purely language-agnostic scaling.

\paragraph{Lesson 1: Reliable Resources Require Language-Specific Effort}
High-quality parallel data collection plays a crucial role for low-resource languages. Carefully curated bitext substantially improves embedding alignment and downstream cross-lingual performance, including cross-lingual retrieval, semantic search, and transfer-based classification. Parallel data acts as an anchoring mechanism: it reinforces semantic correspondence between languages, stabilizes multilingual representations, and reduces drift during training and adaptation.

However, parallel data is also precisely where low-resource settings often fail in practice. While modern bitext mining systems for high-resource language pairs are remarkably strong and can extract parallel sentences reliably at scale, their performance degrades substantially for low-resource languages. This is not only due to limited data volume, but also due to domain mismatch, weaker multilingual encoders for the target language, and higher noise levels in the crawled web \citep{kreutzer-etal-2022-quality}.

\begin{figure}
    \centering
    \includegraphics[width=1.0\linewidth]{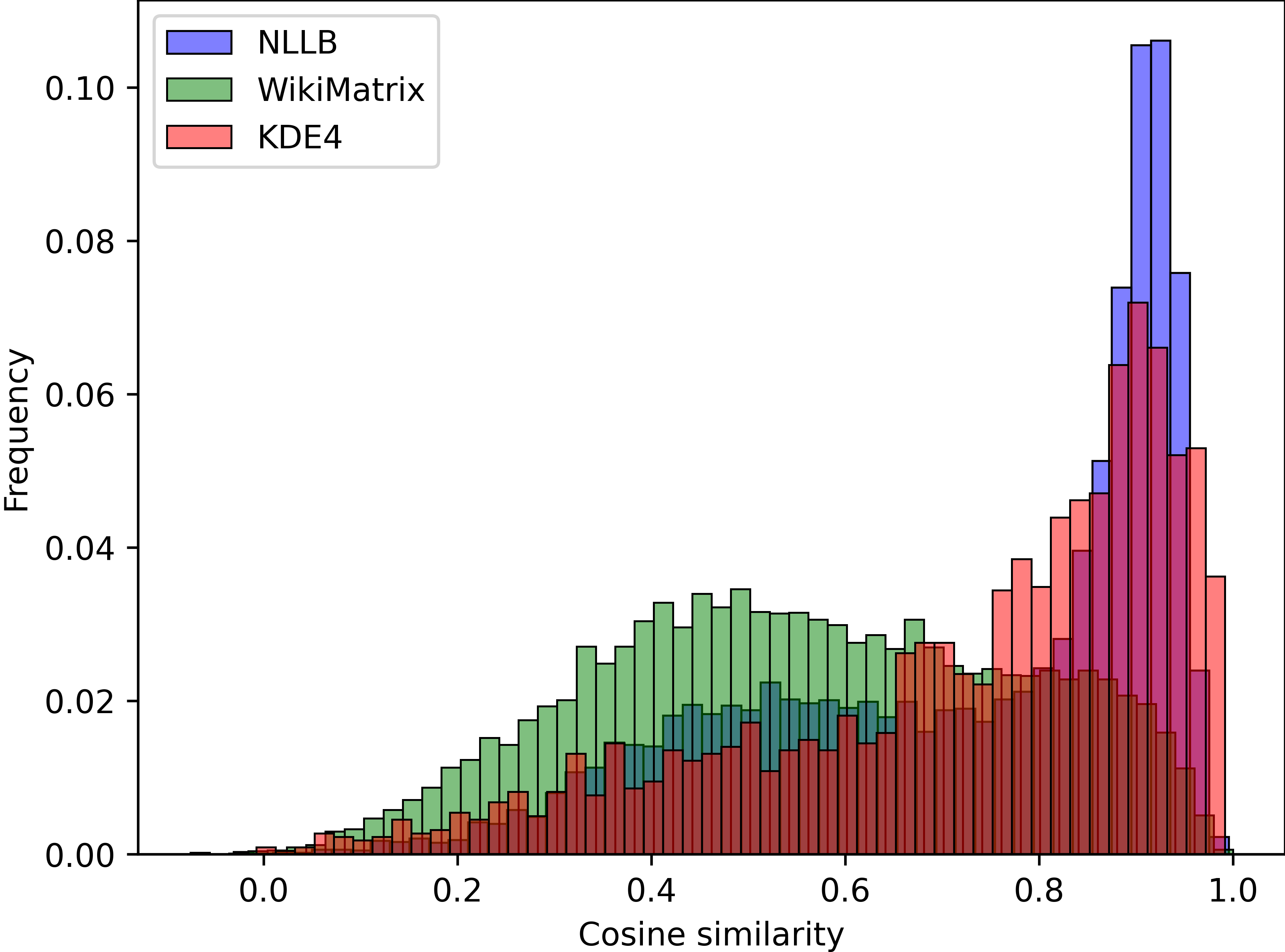}
    \caption{\textbf{Distribution of cosine similarities between EN-LB sentence pairs.} More details are provided in Appendix \ref{app:similarity_histogram}.}
    \label{fig:similarity_histogram}
\end{figure}

Moreover, even when parallel corpora are reported as available, they may be unusable in practice. For Luxembourgish, we find that a non-trivial portion of Luxembourgish segments in widely used parallel datasets are not actually even Luxembourgish (Table \ref{tab:lang_id}), and many mined sentence pairs are only weakly related or entirely unrelated (Figure \ref{fig:similarity_histogram}). This pattern of data degradation is well-documented across the low-resource landscape. Similar quality concerns have been raised regarding Sinhala and Tamil \citep{ranathunga-etal-2024-quality}, Catalan \citep{de-gibert-bonet-etal-2022-quality}, and a wide array of other languages analyzed in the comprehensive audit by \citet{kreutzer-etal-2022-quality}.

This suggests that for low-resource languages, the bottleneck is often not the absence of parallel data per se, but the absence of \emph{reliable} parallel data.

\begin{table}[]
    \centering
    \begin{tabular}{|c|c|}
        \hline Dataset & \% Non-LB Text \\ \hline
        WikiMatrix & 0.77 \\
        NLLB & 21.39 \\
        KDE4 & 70.05 \\ 
        CCMatrix & 99.42  \\ \hline
    \end{tabular}
    \caption{Proportion of EN–LB sentence pairs in which the LB segment was identified as non-LB. For CCMatrix and NLLB, estimates are computed from 100k-sample subsets due to corpus size. More details are provided in Appendix \ref{app:lang_id}.}
    \label{tab:lang_id}
\end{table}

As a result, building usable parallel corpora for low-resource languages frequently requires human intervention, not necessarily in the form of full manual translation, but through targeted guidance and language-aware constraints that improve mining precision.

A concrete example is presented by \citet{philippy-etal-2025-luxembedder}, who construct high-quality Luxembourgish--French and Luxembourgish--English parallel corpora from multilingual news articles published by a Luxembourgish news provider. Although this multilingual data source is publicly available, existing large-scale automatic pipelines have so far been unable to reliably mine parallel corpora for Luxembourgish from it. The difficulty lies in three practical challenges: (1) the three languages (EN, FR \& LB are published under different domains and websites with differing structures, which may break automatic link-based parallel webpage mining pipelines \citep[e.g.,][]{liu-etal-2014-iterative}; (2) not every article is published in all three languages, creating an asymmetry across language pairs; and (3) even when articles report on the same event, they are not always literal translations and may content-wise differ systematically due to target-audience differences.

Consequently, standard language-agnostic bitext crawlers and same-domain URL pairing fail, and full-document matching is unreliable. Instead, \citet{philippy-etal-2025-luxembedder} first perform broad semantic article matching, further restricting candidate pairs to articles published within a three-day window to improve robustness. Only after this coarse alignment step, they proceed to sentence-level matching, using additional constraints such as length-based filtering to reduce noise.

In other words, high-quality parallel data could be created precisely because the mining pipeline incorporated language- and source-specific knowledge, human-provided cues, and structural constraints. This type of targeted intervention is difficult, and in some cases impossible, to replicate with a purely automatic, fully language-agnostic bitext mining pipeline.

\paragraph{Lesson 2: The Complementarity of Language-Specific and Cross-Lingual Signals}
While language-specific efforts are indispensable for developing NLP resources for Luxembourgish, these efforts should not result in linguistic isolation. As seen in lesson 1, low-resource languages require tailored data collection and preprocessing because large ``language-agnostic'' pipelines often fail to capture their structural specificities. However, once such resources are built, it becomes equally important to consider how they connect to the broader multilingual landscape.

\citet{koksal_2025_muri} introduced an automatic, largely language-agnostic pipeline for constructing instruction-tuning data across multiple languages, including Luxembourgish. The core idea is to generate instructions based on existing content in the target language, thereby producing synthetic supervision without requiring large-scale manual annotation. Conceptually, such an approach promises to reduce reliance on high-resource languages and strengthen monolingual supervision.

However, subsequent analysis by \citet{philippy2025luxinstructcrosslingualinstructiontuning} identified several limitations of this approach in the Luxembourgish setting\footnote{These limitations do most likely also apply to many other languages.}. Due to the comparatively weak performance of LLMs in Luxembourgish, automatically generated instructions often suffered from reduced fluency, semantic imprecision, and structural inconsistencies. These issues resulted in a dataset of uneven quality, highlighting that language-agnostic generation pipelines are not immune to representational asymmetries between languages.

To address these limitations, \citet{philippy2025luxinstructcrosslingualinstructiontuning} adapted the pipeline specifically to Luxembourgish by incorporating additional language resources and applying more rigorous filtering and heuristic cleaning procedures. More fundamentally, the instruction generation process was modified and, instead of generating instructions directly in Luxembourgish, instructions were produced in English, French, and German. In this setup, the model only needed to understand Luxembourgish source content, but did not have to generate complex instructional phrasing in Luxembourgish itself. This shift leveraged the model’s stronger generative capabilities in high-resource languages while maintaining Luxembourgish task grounding.

Few-shot prompting experiments showed that demonstration examples containing instructions in English, French, or German, while requiring outputs in Luxembourgish, led to higher performance than demonstrations written entirely in Luxembourgish. In other words, cross-lingual instructions did not merely compensate for weak Luxembourgish generation; they frequently outperformed fully monolingual prompting configurations.

This finding carries an important implication. Language-specific data is essential, but simply localizing all components of the training or prompting pipeline to the target language does not automatically improve results. When underlying multilingual representations are asymmetric, high-resource languages may provide a more stable scaffolding for task reasoning and instruction following. Carefully combining cross-lingual strengths with target-language grounding can therefore yield better performance than purely monolingual approaches.
This phenomenon is further supported by the empirical findings of \citet{chen-etal-2024-monolingual} and \citet{li-etal-2024-x}.

This lesson reinforces the broader thesis of this paper: language-specific efforts are necessary, but they must be integrated with, rather than isolated from, cross-lingual transfer mechanisms.

\paragraph{Lesson 3: It Is Not Only How We Transfer, but What We Transfer}

A traditional and still widely adopted approach to zero-shot topic classification involves fine-tuning a model on Natural Language Inference (NLI) datasets \citep{yin-etal-2019-benchmarking}. In NLI training, the model learns to determine the relationship between a premise and a hypothesis, typically classifying it as entailment, contradiction, or neutral. At inference time, topic classification is reframed within this paradigm: the input text is treated as the premise, and each candidate topic label is formulated as a hypothesis (e.g., ``This text is about politics.''). The model assigns an entailment score to each hypothesis, and the label with the highest entailment probability is selected as the predicted topic.

Given the scarcity of NLI datasets for low-resource languages, this framework is often regarded as particularly suitable for cross-lingual transfer. In practice, a model can be fine-tuned on NLI data in English, or another high-resource language, and subsequently applied directly to a target low-resource language in a zero-shot setting.

However, in practice, this assumption does not always hold for low-resource languages. NLI constitutes a cognitively and linguistically demanding task, as it requires fine-grained semantic understanding, logical reasoning, and sensitivity to subtle pragmatic cues in the language under consideration. For language models operating in low-resource settings, such capabilities may be underdeveloped due to limited pretraining exposure. As a result, fine-tuning on NLI data, whether in the target language itself or transferred from high-resource languages, may fail to yield substantial gains, because the task complexity can exceed the model’s effective linguistic competence in that language. In this context, it may be more beneficial to rely on a comparatively simpler proxy objective that enables the model to acquire at least partial semantic competence, rather than optimizing it for a demanding inference task from which it cannot fully benefit.

This perspective is empirically supported by the findings of \citet{philippy-etal-2024-forget}, who investigated zero-shot topic classification for Luxembourgish. Their experiments demonstrated that the conventional NLI-based paradigm is suboptimal for Luxembourgish, regardless of whether NLI supervision is provided directly in Luxembourgish or transferred from high-resource languages such as German, English, or French. Instead, they introduced an alternative based on a Luxembourgish-specific lexical resource containing synonyms, translations, and example sentences. From this resource, they constructed a training dataset and optimized the model using a contrastive learning objective: given a sentence containing a target word, the model must determine whether a candidate word constitutes a valid synonym or translation. In other words, rather than training the model to perform premise–hypothesis entailment, they directly reinforced Luxembourgish semantic relationships through a more accessible and linguistically grounded signal, which is considerably easier to obtain than large-scale NLI annotations for many under-resourced languages.

More broadly, the results underscore a central principle: cross-lingual transfer is shaped not only by how knowledge is transferred, but by what kind of knowledge is transferred. Effective transfer requires a match between the complexity of the objective and the model’s existing linguistic competence in the target language. For under-resourced languages in particular, carefully designed, linguistically grounded proxy tasks may offer a more reliable path toward robust performance than directly transferring high-level reasoning objectives.

\section{Towards Balancing Transfer and Language-Specific Effort}
Building sustainable NLP systems for low-resource languages requires more than applying cross-lingual transfer at scale. The Luxembourgish case shows that transfer succeeds only when certain linguistic, data, and modeling conditions are met. The following guidelines outline practical recommendations for balancing cross-lingual leverage with targeted language-specific development, treating the two as complementary components of a robust low-resource NLP pipeline.

\paragraph{Treat Transfer as Conditional, Not Automatic}
Cross-lingual transfer should not be treated as a guaranteed by-product of multilingual pretraining. While multilingual models often show strong zero-shot or few-shot capabilities, their ability to generalize across languages depends on conditions that are not uniformly satisfied across languages. Shared parameterization alone does not ensure stable or robust transfer.

Before relying on transfer-based methods, it is therefore essential to assess whether minimal target-side prerequisites are in place. These prerequisites may include sufficient pretraining exposure, coherent subword representations, basic lexical and semantic competence. If these foundational elements are weak or absent, transfer may appear unstable, inconsistent, or task-dependent.

In practical terms, this implies that researchers and practitioners should conduct lightweight diagnostic checks prior to downstream deployment. Examples include inspecting tokenization fragmentation, testing basic sentence similarity performance, or probing for language confusion in multilingual contexts. Such diagnostics help determine whether the model possesses enough internal grounding in the target language for transfer-based approaches to be effective.

In short, cross-lingual transfer depends on underlying conditions rather than being an unconditional property of multilingual models. Verifying these conditions can prevent misattributing transfer failures to model architecture or task difficulty when they are in fact rooted in insufficient target-language grounding.

\paragraph{Assess and Curate Resource Quality Before Use}
The mere availability of large-scale datasets does not guarantee their usefulness in low-resource settings. While multilingual corpora, such as automatically mined parallel datasets, often contain valuable material, they may also include substantial noise, language misidentification, weak semantic alignment, or domain mismatches, all of which tend to affect low-resource languages more severely.

As illustrated in the Luxembourgish case, a non-trivial portion of widely used parallel data may not even be written in the intended target language, and many mined sentence pairs may exhibit only superficial or partial semantic correspondence. Without careful filtering and validation, such noise can undermine representation quality, weaken alignment, and ultimately reduce downstream performance. Therefore, before integrating large-scale resources into a transfer pipeline, it is crucial to evaluate their reliability. A smaller, high-quality dataset may anchor cross-lingual representations more effectively than a large but noisy corpus.

In essence, the question is not whether data exists, but whether it is sufficiently trustworthy to serve as stable supervision.

\paragraph{Leverage High-Resource Languages as Scaffolding}
High-resource languages can serve as valuable scaffolding when developing resources for low-resource settings. Although fully monolingual resources may appear conceptually cleaner, their practical impact depends on the model’s existing competence in the target language. If representation in the underlying LLM is weak, even carefully constructed monolingual datasets may yield limited gains, as the model may struggle to meaningfully interpret or generalize from them.

In such contexts, selectively incorporating high-resource languages can provide stabilizing anchors. Combining low-resource materials with related high-resource content can strengthen semantic alignment and facilitate learning, allowing the model to better contextualize and use the target-language data. Rather than diminishing language-specific efforts, this cross-lingual grounding can enhance their effectiveness within a broader multilingual framework.

\paragraph{Favor Targeted Interventions Over Large-Scale Expansion}
In low-resource settings, improvements often arise from small, targeted interventions rather than large-scale resource expansion. Addressing specific bottlenecks, such as correcting systematic tokenization issues, introducing small curated evaluation sets, or adding narrowly focused training data, can yield disproportionate gains relative to the effort involved. Because low-resource pipelines are particularly sensitive to representation gaps and data noise, incremental improvements guided by careful evaluation can be more effective than simply scaling up data collection. Prioritizing targeted refinements allows to progressively strengthen weak points in the system while maintaining overall stability.
\section{Conclusion}
Cross-lingual transfer is central to extending NLP to low-resource languages. However, our findings show that transfer alone rarely yields robust systems. Using Luxembourgish as a case study, we demonstrate that even languages with favorable conditions, including typological proximity to high-resource languages, a shared script, and a relatively strong digital presence, still face substantial limitations when relying solely on cross-lingual transfer.

We observe that cross-lingual transfer is most effective when supported by targeted language-specific efforts. High-quality parallel data and linguistically grounded supervision help stabilize multilingual representations and enable meaningful transfer. Conversely, purely language-specific approaches are rarely sufficient, since low-resource settings typically lack enough data to achieve strong performance without cross-lingual signals.
Cross-lingual transfer and language-specific development are therefore best understood as complementary strategies. Language-specific resources ground the modeling of the target language, while cross-lingual transfer allows these resources to benefit from the broader capacity of multilingual models.

More broadly, the Luxembourgish case highlights an important implication for multilingual NLP. Structural proximity to high-resource languages does not guarantee strong cross-lingual performance. If limitations appear even under such favorable conditions, they are likely to be even more pronounced for languages that are typologically distant, digitally underrepresented, or socioeconomically marginalized.

Ultimately, advancing NLP for the majority of the world's languages requires moving beyond the assumption that cross-lingual transfer alone can close the resource gap. Sustainable progress instead depends on integrating cross-lingual modeling with deliberate language-specific resource development.

\section*{Limitations}
This work assumes that Luxembourgish approximates an upper bound for cross-lingual transfer among low-resource languages. This assumption is partly motivated by empirically studied factors known to facilitate transfer, such as linguistic similarity and lexical overlap with high-resource languages. At the same time, it also relies on additional contextual properties, such as the institutionalization of the language or its relatively strong digital presence, that are intuitively beneficial but have received less systematic empirical investigation in multilingual NLP.

As a result, the upper-bound framing should be interpreted as a theoretical approximation rather than a strictly validated claim. While these properties plausibly create favorable conditions for language technology development, they are not necessarily fully predictive of model performance. Multilingual models may also be influenced by less visible or difficult-to-measure factors, such as properties of pretraining data or other aspects of training pipelines. Consequently, although Luxembourgish exhibits many characteristics that are advantageous compared to most low-resource languages, we do not exclude the possibility that other languages with similar conditions may exist.

\bibliography{latex/custom, latex/anthology-1-reduced}

\appendix

\section{Details About Figures and Tables}

\subsection{Figure \ref{fig:lang_dist}} \label{app:lang_dist}
Figure \ref{fig:lang_dist} visualizes a two-dimensional Principal Component Analysis (PCA) projection of language representations constructed by concatenating syntactic, phonological, inventory, genetic (language family), and geographical feature vectors obtained from \textit{lang2vec} \citep{littell-etal-2017-uriel}. For syntactic, phonological, and inventory features, we relied on the KNN-based representations to guarantee vectors of consistent dimensionality across languages. The resulting vectors were standardized and reduced to two dimensions using PCA. Language-level resource availability was operationalized as the number of Wikipedia articles per language, extracted from the Wikimedia statistics page\footnote{\url{https://meta.wikimedia.org/wiki/List_of_Wikipedias} visited on March 3, 2026}. In the visualization, each point corresponds to a language positioned according to its first two principal components, and colored by the logarithm of its Wikipedia article count (log(n+1)). Luxembourgish (lb) is highlighted in black for reference.

\subsection{Figure \ref{fig:speakers_vs_articles}} \label{app:speakers_vs_articles}
The estimated number of speakers per language was obtained from LinguaMeta \citep{ritchie-etal-2024-linguameta}.

For Wikipedia, the number of articles per language edition was collected from the Wikimedia statistics page\footnote{\url{https://meta.wikimedia.org/wiki/List_of_Wikipedias} visited on March 3, 2026}.

For Common Crawl, we used language-level page counts extracted from the CC-MAIN-2026-04 crawl \footnote{\url{https://commoncrawl.github.io/cc-crawl-statistics/plots/languages}}.

\subsection{Table \ref{tab:lang_id}} \label{app:lang_id}
To estimate the proportion of non-Luxembourgish segments, we apply automatic language identification to the Luxembourgish side of each English–Luxembourgish sentence pair. We use OpenLID-v3 \citep{fedorova2026openlidv3improvingprecisionclosely} with a threshold of 0.5, which we found to perform reliably for Luxembourgish in preliminary experiments. We evaluate the largest English–Luxembourgish corpora listed in OPUS\footnote{\url{https://opus.nlpl.eu/corpora-search/en&lb}} \citep{TIEDEMANN12.463}: WikiMatrix \citep{schwenk-etal-2021-wikimatrix}, CCMatrix \citep{schwenk-etal-2021-ccmatrix}, NLLB \citep{nllbteam2022languageleftbehindscaling}, and KDE4\footnote{\url{https://huggingface.co/datasets/Helsinki-NLP/kde4}}, sampling 100,000 sentence pairs from CCMatrix and NLLB due to their size.

\subsection{Figure \ref{fig:similarity_histogram}} \label{app:similarity_histogram}

We first remove all segments predicted as non-Luxembourgish by OpenLID-v3 \citep{fedorova2026openlidv3improvingprecisionclosely} from NLLB \citep{nllbteam2022languageleftbehindscaling}, WikiMatrix \citep{schwenk-etal-2021-wikimatrix}, and KDE4. From each filtered dataset, we then select up to 10,000 English–Luxembourgish sentence pairs (using all available pairs in cases where fewer remain). For this subset, we compute sentence embeddings with LaBSE\footnote{\url{https://huggingface.co/sentence-transformers/LaBSE}} \citep{feng-etal-2022-language} and calculate the cosine similarity for each aligned pair. The resulting similarity scores are visualized as normalized histograms.

\end{document}